\documentclass{article}

\PassOptionsToPackage{numbers, compress}{natbib}
\usepackage[preprint]{neurips_2021}

\usepackage[utf8]{inputenc} % allow utf-8 input
\usepackage[T1]{fontenc}    % use 8-bit T1 fonts
\usepackage{hyperref}       % hyperlinks
\usepackage{url}            % simple URL typesetting
\usepackage{booktabs}       % professional-quality tables
\usepackage{amsfonts}       % blackboard math symbols
\usepackage{nicefrac}       % compact symbols for 1/2, etc.
\usepackage{microtype}      % microtypography
\usepackage{amsmath}
\usepackage{graphicx}
\usepackage{xcolor}         % colors
\usepackage{float}

\usepackage[hang,flushmargin]{footmisc}

\usepackage{sidecap}
\sidecaptionvpos{figure}{t}

\usepackage[inline]{enumitem}

\graphicspath{ {./figs/} }

\newcommand{\figscaleB}{0.6}

\title{More Than Meets the Eye: Self-Supervised Depth Reconstruction From Brain Activity}

\author{
    Guy Gaziv \hspace{3.54cm} Michal Irani \\
    Dept. of Computer Science and Applied Math \\
    The Weizmann Institute of Science \\
}
\begin{document}

\maketitle
\vspace*{-0.3cm}
% \vspace*{-0.4cm}
\begin{abstract}
% \vspace*{-0.1cm}

%%%%%%%%%%%%%%%%%%%%%%%%%%%%%% TL;DR %%%%%%%%%%%%%%%%%%%%%%%%%%%%%%%%%
% We propose a method for decoding the 3D depth maps of 2D natural images that were observed by human subjects directly from their fMRI brain recordings.

% A method for decoding dense 3D depth of observed natural images directly from fMRI brain recordings.
%%%%%%%%%%%%%%%%%%%%%%%%%%%%%%%%%%%%%%%%%%%%%%%%%%%%%%%%%%%%%%%%%%%%%%

%%%%%%%%%%%%%%%%%%%%%%%%%%%%%% KEYWORDS %%%%%%%%%%%%%%%%%%%%%%%%%%%%%%%%%
% fMRI decoding, Depth Reconstruction, Image Reconstruction, Self-Supervised Learning, Computer Vision.
%%%%Self-Supervised Learning, Decoding, Encoding, fMRI, Image Reconstruction, Vision
%%%%%%%%%%%%%%%%%%%%%%%%%%%%%%%%%%%%%%%%%%%%%%%%%%%%%%%%%%%%%%%%%%%%%%

% Guy May28
In the past few years, significant advancements were made in reconstruction of observed natural images from fMRI brain recordings using deep-learning tools. Here, for the first time, we show that dense 3D depth maps of observed 2D natural images can also be recovered directly from fMRI brain recordings. We use an off-the-shelf method to estimate the unknown depth maps of natural images. This is applied to both: (i)~the small number of images presented to subjects in an fMRI scanner (images for which we have fMRI recordings – referred to as ``paired'' data), and (ii)~a very large number of natural images with no fMRI recordings (``unpaired data''). The estimated depth maps are then used as an auxiliary reconstruction criterion to train for depth reconstruction directly from fMRI. We propose two main approaches: Depth-only recovery and joint image-depth RGB\textbf{D} recovery. Because the number of available ``paired'' training data (images with fMRI) is small, we enrich the training data via self-supervised cycle-consistent training on many ``unpaired'' data (natural images \& depth maps without fMRI). This is achieved using our newly defined and trained Depth-based Perceptual Similarity metric as a reconstruction criterion. We show that predicting the depth map \textit{directly from fMRI} outperforms its indirect sequential recovery from the reconstructed images. We further show that activations from early cortical visual areas dominate our depth reconstruction results, and propose means to characterize fMRI voxels by their degree of depth-information tuning. This work adds an important layer of decoded information, extending the current envelope of visual brain decoding capabilities.

\end{abstract}

%   The abstract paragraph should be indented \nicefrac{1}{2}~inch (3~picas) on
%   both the left- and right-hand margins. Use 10~point type, with a vertical
%   spacing (leading) of 11~points.  The word \textbf{Abstract} must be centered,
%   bold, and in point size 12. Two line spaces precede the abstract. The abstract
%   must be limited to one paragraph.
\vspace*{-.2cm}
\begin{center}
    {\small
    \emph{\textbf{Code:}} \href{https://github.com/WeizmannVision/SelfSuperReconst}{\color{magenta}{\texttt{https://github.com/WeizmannVision/SelfSuperReconst}}}
    }
\end{center} 
\vspace*{-.1cm}
\section{Introduction}
% \vspace*{-0.2cm}

% \begin{figure}[t] 
% \centering
% % \includegraphics[scale=\figscale]{TaskMethod3.pdf}
% % \includegraphics[scale=\figscalehalf]{Idea.pdf}
% \includegraphics[scale=\figscale]{Idea.pdf}
% \caption{
% \textbf{Our proposed method.}
% {\small {\it \textbf{(a)} The task: reconstructing color images (RGB) \& dense depth maps (D) from evoked brain activity, captured via fMRI. \textbf{(b)} Predicting depth maps for all ground truth images (RGB) using a designated pretrained network (called ``MiDaS''). This includes the images in the paired fMRI dataset and those in the external dataset of unpaired images. \textbf{(b), (c)} Supervised training for decoding (\textbf{b}) and encoding (\textbf{c}) using limited training pairs. This gives rise to poor generalization. \textbf{(d), (e)} Illustration of our added self-supervision, which enables training on ``unlabeled images'' (any natural image with no fMRI recording -- (\textbf{d})), and on the ``unlabeled fMRI'' (fMRI data without any corresponding images -- (\textbf{e})). In particular, the latter allows adapting the decoder to the statistics of the target test-fMRI despite not having any information about their corresponding images.}}
% }
% \label{fig:Idea}
% \vspace*{-0.3cm}
% \end{figure}

\begin{figure}[t] 
\vspace*{-0.5cm}
% \begin{SCfigure}
\centering
\includegraphics[scale=0.42]{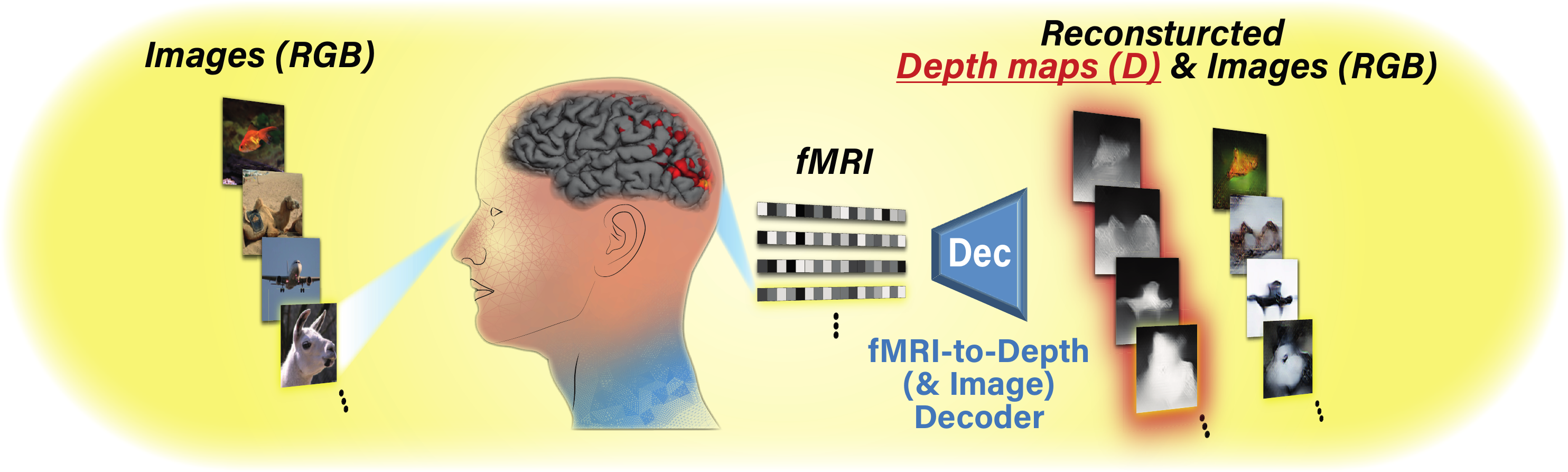}
\vspace*{-.1cm}
\caption{
% \textbf{The task: reconstructing color images (RGB) \& dense depth maps (D) from evoked brain activity captured via fMRI.}
\textbf{The task: reconstructing dense depth maps (D) from fMRI brain recordings.}
}
\label{fig:Task}
% \vspace*{-0.5cm}
\vspace*{-1cm}
% \vspace*{-.6cm}
% \vspace*{-0.91cm}
\end{figure}

Decoding observed visual scene information from brain activity may form the basis for brain-machine interfaces and for understanding visual processing in the brain (Fig \ref{fig:Task}). A classic challenge in this domain is reconstructing seen natural images from their recorded fMRI\footnote{functional magnetic resonance imaging.} brain activity~\cite{Kamitani2005DecodingBrain, Haynes2006DecodingHumans, Kay2009ISee, Naselaris2009BayesianActivity}. 
%The goal in this task 
%
To learn such  mappings, fMRI datasets provide pairs of images and their corresponding fMRI responses, referred to here as ``paired'' data. 
The goal in that challenge is to learn  fMRI-to-image decoding which generalizes well to image reconstruction from novel ``test-fMRIs'' induced by novel images.
% Their goal is to learn  fMRI-to-image decoding which generalizes well to image reconstruction from novel ``test-fMRIs'' induced by novel images.
%(referred to as ``test-images''). 
%

% Humans perform well at monocular depth estimation by
% exploiting cues such as perspective, scaling relative to the known size of familiar objects, appearance in the form of lighting and shading and occlusion~\cite{Howard2012PerceivingDepth}. This combination of both top-down and bottom-up cues appears to link full scene understanding with our ability to accurately estimate depth.

However visual scene understanding goes well beyond the RGB bitmap which represents it. An important complementary cue to natural-image understanding is inferring depth relations within it~\cite{Iyer2018DepthScenes}. 
Humans perform well on monocular depth estimation by
exploiting cues such as perspective, scaling relative to the known size of familiar objects, shading and occlusion~\cite{Howard2012PerceivingDepth}.
Furthermore, previous studies found evidence for depth cue encoding and integration in the human visual cortex~\cite{Welchman2016The3D, Armendariz2019ArealHuman}. 

In this paper propose a new challenge, which goes beyond the traditional fMRI-to-image reconstruction task: \textbf{\textit{Given an fMRI recording of an  observed scene (a 2D image), reconstruct its underlying dense 
3D depth map (D) directly from fMRI.}} This can be done in addition to (or without) recovering the observed RGB image.
Adding this additional layer of depth information to the fMRI decoding problem has two potential implications: (i)~It paves the way to reconstructing \textit{new types of inferred dense information about a scene that is not explicitly presented to the subject}, and (ii)~it provides auxiliary criteria to guide and train image-reconstruction networks, which complements existing reconstruction criteria based purely on RGB data.

% We use an off-the-shelf method to estimate the unknown depth maps of the presented images. We then use the estimated maps as an auxiliary reconstruction criterion within our self-supervised framework for image reconstruction: predicting an RGB\textbf{D} output for a given fMRI input. We show that predicting the depth map \textit{directly from fMRI} outperforms its recovery from a reconstructed image. Beyond introducing this new capability, we demonstrate that adding depth recovery criteria gives rise also to better image (RGB) reconstruction and semantic classification results than those obtained without it.

% \emph{However, the shortage of ``paired'' training data limits the generalization power of today's fMRI decoders}.
% The number of obtainable image-fMRI pairs is bounded by the limited time a human can spend in an MRI scanner. Accordingly, most datasets provide only up to a few thousands of (Image,fMRI) pairs.
% Such limited data cannot span the huge space of natural images. 
% % Moreover, the poor spatio-temporal resolution of fMRI signals, as well as their low Signal-to-Noise Ratio (SNR), reduce the reliability of the already scarce paired training data~\cite{Lage-Castellanos2019MethodsResponses, Glover2011OverviewImaging, Glasser2013TheProject}.

\textit{\textbf{Prior work on image reconstruction from fMRI.}}
%Methods for image reconstruction from fMRI can broadly be classified into three main categories: 
fMRI-to-Image reconstruction methods can broadly be classified into three main categories: 
\begin{enumerate*}[label=(\roman*)] 
    \item Linear regression between fMRI data and handcrafted image-features (e.g., Gabor wavelets)~\cite{Kay2008IdentifyingActivity, Naselaris2009BayesianActivity, Nishimoto2011ReconstructingMovies.},
    \item Linear regression between fMRI data and pre-trained deep image-features -- e.g., features of pretrained AlexNet~\cite{Wen2018NeuralVision, Guclu2015DeepStream, Shen2019DeepActivity, Zhang2018Constraint-FreeNetwork}, or latent spaces of pretrained generative models~\cite{Han2019VariationalCortex, Ren2021ReconstructingLearning, MozafariReconstructingBigBiGAN, Qiao2020BigGAN-basedActivity}, and
    \item End-to-end Deep Learning~\cite{Shen2019End-to-endActivity, St-Yves2019GenerativeImages, Seeliger2018GenerativeActivity, Lin2019DCNN-GAN:FMRI, Gaziv2020, Beliy2019FromFMRI}. 
\end{enumerate*}

Most of these methods inherently rely on ``paired'' data to train their decoder (pairs of images and their corresponding fMRI responses). 
In the typical case, when only a small number of such pairs are available, purely supervised models are prone to overfitting. This leads to poor generalization to new test-data (fMRI response evoked by new images). 
Recently, \cite{Gaziv2020, Beliy2019FromFMRI}  proposed to cope with the limited ``paired'' training examples by adding self-supervision on additional ``unpaired'' natural images (images with no fMRI recording). This led to state-of-the-art results in image-reconstruction. 
However, all the above methods focused  on reconstructing only images and semantic features, glossing over other important visual perception cues, such as scene depth. This is the focus of  the current paper.

\begin{figure}[t] 
\vspace*{-.5cm}
\hspace*{-1cm}
\centering
\includegraphics[scale=.42]{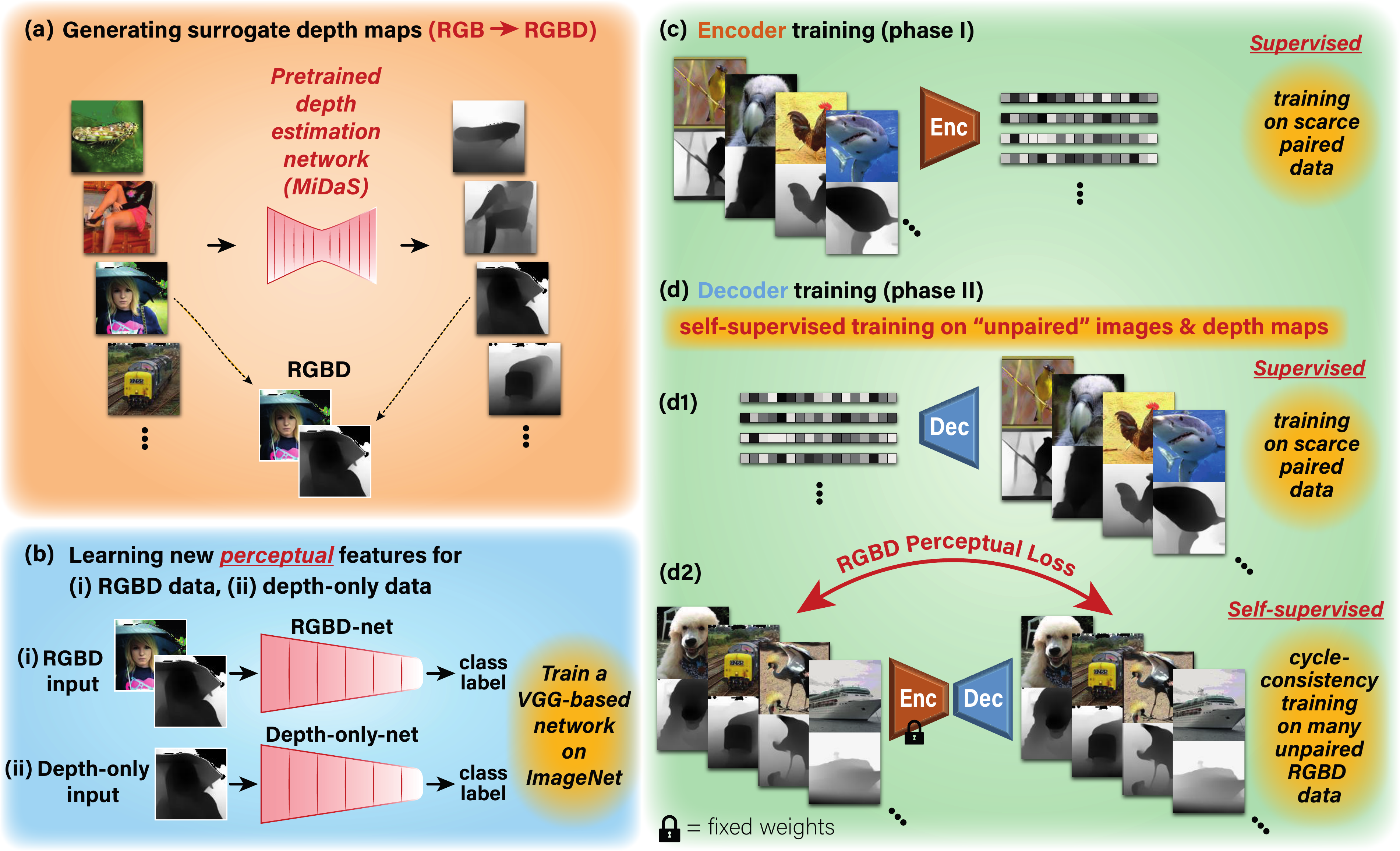}
\vspace*{-.5cm}
\caption{
\textbf{Our proposed method.}
{\small {\it \textbf{(a)}~Predicting depth maps for all ground truth images (RGB) using a pretrained network (``MiDaS''~\cite{Lasinger2019TowardsTransfer}). This includes the images in the paired fMRI dataset and those in the external dataset of unpaired natural images. \textbf{(b)}~Learning depth perceptual features. We train two VGG-based networks for ImageNet object recognition for either input type, RGBD or Depth-only. These networks give rise to new types of perceptual metrics for Decoder training and provide the Encoder backbone. \textbf{(c)}~Phase I: Supervised training of the Encoder with ``paired'' training data. \textbf{(d)}~Phase II: Training the Decoder with two types of data simultaneously: \textbf{(d1)}~The ``paired'' training data (supervised examples), and \textbf{(d2)}~``unpaired'' natural images with their depth maps (self-supervision). 
% {Note that the test-\underline{images} are never used for training.} 
The pretrained Encoder from phase I is kept fixed in phase II.
}}
}
\label{fig:Method}
% \vspace*{-0.3cm}
\vspace*{-0.45cm}
\end{figure}

\begin{figure} 
\vspace*{-1cm}
\hspace*{-1cm}
% \begin{SCfigure}
    % \includegraphics[scale=\figscaleC]{StarterResults2.pdf}
% \includegraphics[scale=\figscaleB]{RGBD_Reconst.pdf}
% \includegraphics[scale=\figscaleB]{RGBD_Reconst_noDE.pdf}
% \includegraphics[scale=\figscaleB]{RGBD_Reconst2.pdf}
\includegraphics[scale=\figscaleB]{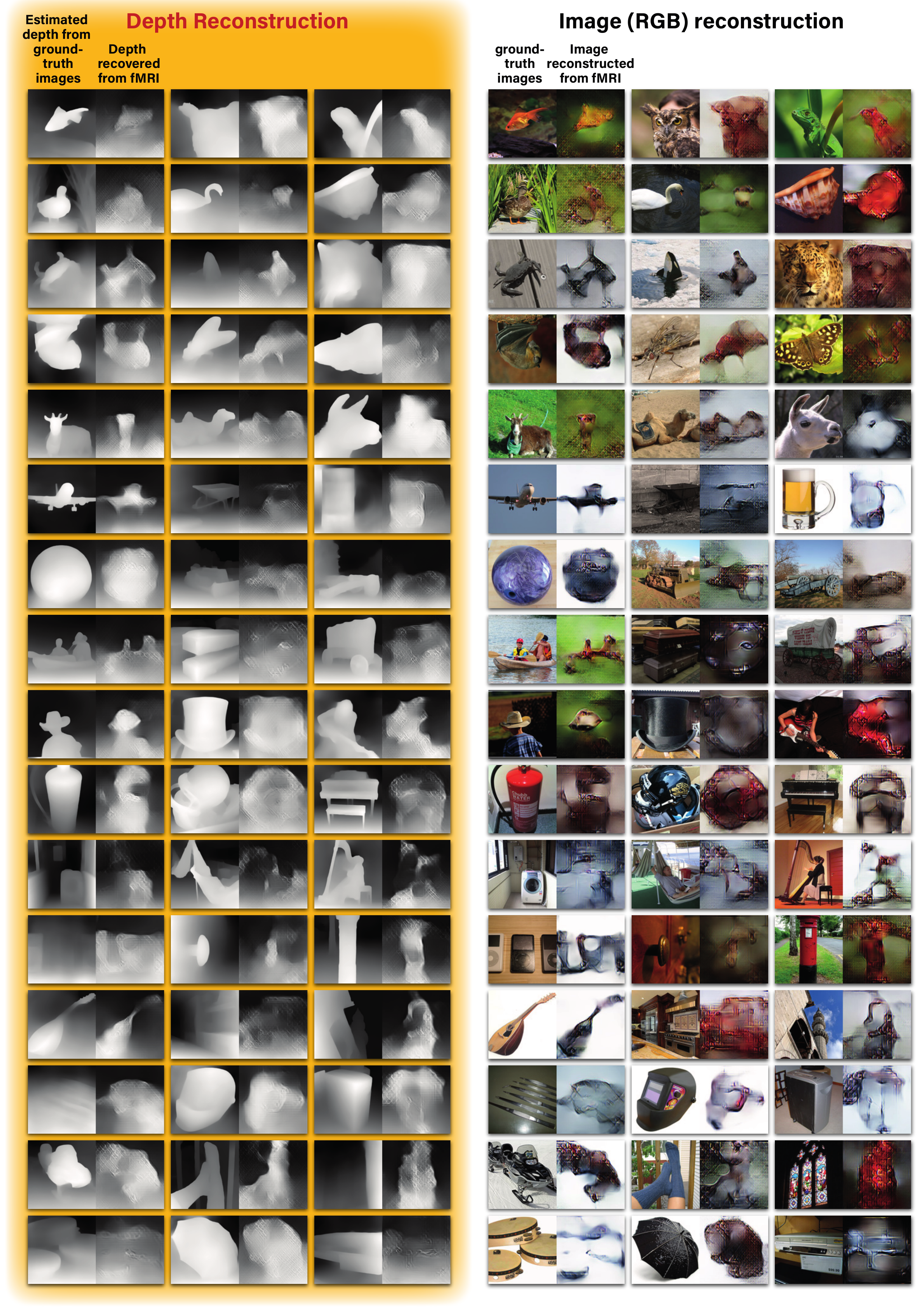}
% \vspace*{-0.7cm}
\vspace*{-0.5cm}
\caption{
\textbf{Image \& Depth (RGBD) reconstruction results.} 
    {\small {\it 
    \textbf{(Left column):} Depth maps reconstruction results side-by-side their estimated ground-truth from the original RGB images. 
    \textbf{(Right column):} Image (RGB) reconstruction results side-by-side the ground-truth images presented to the human. 
    % We show results of the entire test cohort of
    We show results of test cohort of \mbox{\href{https://openneuro.org/datasets/ds001246/versions/1.0.1}{\color{magenta}{fMRI on ImageNet}}~\cite{Horikawa2015GenericFeatures}}. 
    }}}
    \label{fig:MainResults}
    \vspace*{-0.3cm}
    % \vspace*{-0.585cm}
\end{figure}

We present a new approach that generalizes the self-supervised approach of~\cite{Gaziv2020, Beliy2019FromFMRI} to accommodate for \textit{recovery of dense depth maps directly from fMRI}.
Our approach is illustrated in Fig~\ref{fig:Method}:
\newline \ 
$\bullet$
We first estimate the depth maps of natural images using an off-the-shelf pretrained network (``MiDaS''~\cite{Lasinger2019TowardsTransfer}) for monocular depth estimation (Fig~\ref{fig:Method}a).  This is applied both to  the scarce ``paired''  images in the fMRI dataset, as well as to 
% all the ``unpaired'' natural images from ImageNet challenge. This step provides us with
many more ``unpaired'' natural images from ImageNet. This step provides us with
surrogate ``ground-truth''Depth information, which can either be used on its own for training our network, or can be combined with the source image to provide new  RGBD ``ground-truth'' data for training.  
%%%%%%%%%%%%%%%%%%%%%%%%%%%%%%%%%
% START FROM HERE TOMORROW:
%%%%%%%%%%%%%%%%%%%%%%%%%%%%%%%%%
\newline \ 
$\bullet$
We then train two types of deep networks: (i)~an Encoder \emph{Enc}, that encodes depth-based information (either Depth alone, or RGBD data) into their corresponding fMRI responses (Fig~\ref{fig:Method}c), and (ii)~a Decoder \emph{Dec}, that decodes fMRI recordings to their corresponding depth-based information (Fig~\ref{fig:Method}d1). Concatenating those two networks back-to-back, \emph{Enc-Dec}, yields a combined network whose input and output are the same depth-based information (Fig.~\ref{fig:Method}c2). \emph{\textbf{This allows for unsupervised training on \underline{unpaired data}}} (i.e., Depth maps or RGBD data \emph{without fMRI recordings}, e.g., obtained from 50,000 randomly sampled natural images from ImageNet in our experiments). Such self-supervision adapts the network to the statistics of never-before-seen depth-based data. 
\newline \ 
$\bullet$
The loss enforced on the reconstructed depth-based information employs a special \emph{\textbf{Depth-based Perceptual Similarity}}, which we also present in this paper (Fig~\ref{fig:Method}b). This encourages our reconstructed depth maps to be \emph{perceptually meaningful}.
Fig~\ref{fig:MainResults} shows image \& depth reconstructions using our RGBD-based approach.
%combined with surrogate depth data. 
These results demonstrate a new capability of dense Depth recovery directly from fMRI (in addition to  RGB image reconstruction). 
We show that training our networks on Depth-only data or on RGBD data provide comparable quality of depth-map reconstructions from fMRI (Fig~\ref{fig:DepthRecoveryOptions}bc).
We further show that depth reconstructions \emph{directly from fMRI} provide significantly better results than \emph{indirect depth estimation} applied (after the fact) to purely reconstructed RGB-images.
%(see Sec.~\ref{sec:experiments}). 
%
%Importantly, the presented RGB-images were delivered equally to both of the subject's eyes with no binocular disparity.
Lastly, training our networks on the combined RGBD data further allows us to explore whether there are fMRI voxels more tuned to depth information, versus voxels more tuned to RGB information.
These experiments are discussed in Sec.~\ref{sec:experiments}.

\underline{Our contributions are therefore several-fold:}
% \begin{itemize}[noitemsep,nolistsep,leftmargin=*]
\newline \ 
$\bullet$
The first method to reconstruct dense 3D  depth information from brain activity.
\newline \ 
$\bullet$
A self-supervised approach for reconstructing depth-based information (\emph{Depth-only}  or \emph{RGBD}) directly from fMRI recordings, despite having only scarce fMRI training data. 
%To the best of our knowledge, we are the first to suggest an approach for dense depth reconstruction from brain activity.
\newline \ 
$\bullet$
A \emph{Depth-based Perceptual Similarity} measure, based on specially trained \emph{perceptual depth features}.
%We demonstrate high-quality image \& depth map (RGBD) reconstruction from fMRI on a popular fMRI dataset (`fMRI~on~ImageNet'~\cite{Horikawa2015GenericFeatures}).
\newline \ 
$\bullet$
Characterize brain-voxels by their degree of depth-sensitivity via a novel depth-sensitivity measure.

% \end{SCfigure}

\vspace*{-0.2cm}
\section{Overview of the approach} \label{ApproachOverview}
\label{sec:approach}
\vspace*{-0.2cm}
% \vspace*{-0.25cm}
% To train a decoder for reconstruction of images and depth maps, we combine our recently proposed self-supervised method for image reconstruction~\cite{Beliy2019FromFMRI, Gaziv2020} together with off-the-shelf monocular depth estimation method for natural images, called MiDaS~\cite{Lasinger2019TowardsTransfer}. 

Our goal is to reconstruct dense depth 3D information from fMRI data of observed 2D color images. For that purpose we explored three main approaches: (i)~Decoding depth-only information  \emph{directly from fMRI}; (ii)~Simultaneous decoding of images+depth information (RGBD) \emph{directly from fMRI}; (iii)~\emph{Indirect} depth reconstruction computed (after the fact) from RGB images which were reconstructed from fMRI. Our experimental results show (see Sec.~\ref{sec:experiments})  that reconstructing depth \underline{directly from fMRI} (approaches (i) \& (ii)) are significantly superior to sequential \underline{indirect} depth reconstruction (approach~(iii)). 
Fig~\ref{fig:Method} shows our  proposed framework for direct depth reconstruction from fMRI. 
 It consists of three main components: 

% \begin{enumerate}[noitemsep,nolistsep,leftmargin=*]

\textbf{1. \textit{Generating surrogate depth data (Fig~\ref{fig:Method}a).}} \ 
Our training of a depth decoder requires the underlying ground-truth depth maps of natural images.
%: those in the fMRI mapping set, as well as those comprising the external ``unpaired'' images dataset. 
To obtain these data we generate depth maps by feeding-forward the natural images through 
%a designated pretrained network for depth estimation, called MiDaS. 
an off-the-shelf monocular depth estimation method for natural images, called MiDaS~\cite{Lasinger2019TowardsTransfer}.
We then use these maps as our surrogate ``ground-truth'' depth data. We estimated the depth maps of all 1250 images of `fMRI on ImageNet'~\cite{Horikawa2015GenericFeatures}, and all those of ImageNet classification challenge~(ILSVRC)~\cite{Deng2009ImageNet:Database}.
These depth-maps are then used in the following steps. 

\textbf{2. \mbox{\textit{Learning semantically-meaningful depth  features for depth-based perceptual similarity (Fig~\ref{fig:Method}b).}}} \ 
Our trained Decoder (fMRI $\mapsto$ Depth, or fMRI $\mapsto$ RGBD) requires a similarity score/loss on the reconstructed depth maps.  Perceptual Similarity~\cite{Zhang2018TheMetric} has been shown to be a powerful metric for many image reconstruction tasks. However, that metric was developed for images only. We propose here a new perceptually/semantically meaningful similarity score for depth maps. This is obtained by training from scratch two VGG-like networks, trained on  ImageNet for the task of \underline{object recognition}, but from depth-based inputs: (i)~An RGBD network that receives a 4-channel RGBD input (an image concatenated channel-wise with its estimated depth map), and (ii)~A Depth-only network that is trained similarly and with the same architecture, but with a single-channel input containing only the depth map of an image. The goal of such training is not to improve object recognition, but rather to obtain a coarse-to-fine hierarchy of semantically-meaningful depth-related features (depth-only features, or RGBD features). Once trained, these features form the basis to our RGBD and Depth-only Perceptual Similarity metrics used for training and evaluating our reconstructions. 
%, which are used both as a reconstruction criterion and for evaluation of reconstruction quality.  
%These features form the basis of our depth-based perceptual similarity score.
%These RGBD and Depth-only Perceptual Similarity 
%are analogous to the original Perceptual Similarity metric~\cite{Zhang2018TheMetric}, which was also critical to the reconstruction quality in~\cite{Gaziv2020}. However here, they further leverage hierarchical and semantic feature representation that is extracted \textit{\textbf{from the depth data also}}. Lastly, these networks additionally provide the backbone network for the Encoder. 

\textbf{3. \textit{Handling the insufficient fMRI training data (Fig~\ref{fig:Method}cd).}} \ 
To train our Decoder (\mbox{fMRI $\mapsto$ Depth}, or \mbox{fMRI $\mapsto$ RGBD}), we used a moderate-size fMRI dataset --  `fMRI on ImageNet'~\cite{Horikawa2015GenericFeatures}. However, the train-set of this dataset contains only 1200 pairs of images with their corresponding fMRI recordings (this small number is typical of fMRI datasets). Such a small number of examples cannot span the huge space of natural images (nor their corresponding depth maps), resulting in a poor generalization of the Decoder to fMRIs of never-before-seen images. 
% To enrich the training and obtain superior generalization capabilities, we follow the self-supervised approach of~\cite{Gaziv2020}. 
%
To overcome this problem, we generalize the self-supervised approach of~\cite{Gaziv2020} to accommodate for depth information. 
This allows us to train our decoder on many (50,000) additional natural images with their estimated depth maps -- images for which there are \emph{no fMRI recordings}. 
% We thus combine the assets from the previous two steps -- the surrogate depth data and the depth-perceptual feature extractors -- with the self-supervised training approach to allow for self-supervised depth reconstruction from fMRI.
%
Specifically, our semi-supervised approach employs a two-phase training: The first phase is supervised, and focuses on training of an Encoder, $Enc$ -- to map images and their depth maps to their corresponding fMRI recordings. This is done using the 1200 ``paired''  examples from the fMRI dataset, along with their MiDaS-estimated depth maps (Fig~\ref{fig:Method}c). In the second phase, we train the Decoder, $Dec$, using two objectives: (i)~supervised training, to map the limited ``paired'' fMRI recordings to their corresponding images and depth maps (Fig~\ref{fig:Method}d1), and (ii)~\emph{\underline{self-supervised} cycle-consistent training} on a very large number of ``unpaired'' natural images with their corresponding depth maps. This cycle-consistency is facilitated using the auxiliary pretrained Encoder from the first phase (Fig~\ref{fig:Method}d2).
The Decoder loss uses our above-mentioned \emph{depth-based Perceptual Similarity} between the reconstructed and the ``ground-truth'' depth maps. Using this metric ensures that the reconstructed images and their reconstructed depth maps are perceptually and semantically meaningful (well beyond their pixel-level similarity).
%

% \end{enumerate}

\underline{We experimented with 2 main approaches to explore the best scheme for depth-decoding from fMRI:}

\textit{\textbf{Depth-only framework.}} \ 
Our goal is to recover depth. In this mode,  the input to the Encoder \& output of the Decoder (Fig~\ref{fig:Method}cd) is a single-channel depth map; all RGB image data is discarded in the training.  
The reconstruction loss on the decoder in this case is our Depth-only Perceptual Similarity. 
%This approach solely addresses the mapping between depth-maps and corresponding fMRI without knowing anything about the underlying RGB data. 
This approach allowed us to isolate the task of depth recovery from fMRI alone. 
However, it risks poor  Encoder-training in the first supervised phase. The reason being:  there are likely many voxels in the visual cortex that are \emph{not} depth-dependent, only RGB-dependent. Yet, the encoder is expected to predict their values from depth-only information in the first training phase.  
%does not benefit from additional guiding cues for the task.

\textit{\textbf{RGBD framework.}} \ To avoid the above potential problem, we experimented also with training both the Encoder \& Decoder  on combined RGB-D data. The reconstruction loss on the decoder in this case is our RGBD Perceptual Similarity.
This combined reconstruction further allows us to explore whether there are fMRI voxels more tuned to depth, versus voxels more tuned to RGB information. These experiments are discussed in Sec.~\ref{sec:experiments}.

\vspace*{-0.2cm}
\section{Method Details}
\vspace*{-0.2cm}
% \vspace*{-0.25cm}

%Here we detail the specifics of our method. 
%
% We start by describing how we generate surrogate depth data and obtain the depth-perceptual feature extractors. 

% \subsection{Generating surrogate depth data} 
% \label{DepthEstim}
% To obtain the depth maps of the natural images that we train on, we used ``MiDaS'' state-of-the-art method for monocular depth estimation~\cite{Lasinger2019TowardsTransfer}. For best depth estimates we used their non-optimized large model. Due to scale-ambiguity of depth we only considered the non-metric relative differences within an image, i.e., normalized the scale and shift of each depth map using its own internal statistics. For the images in the fMRI paired dataset, we estimated the maps using image resolution of 224$\times$224 and subsequently downscaled to match the Encoder/Decoder working resolution. For the external unpaired images, we estimated the maps using their original resolution and let the same augmentation rules of their RGB counterparts to be applied to them (Resize + CenterCrop).

\underline{\textit{\textbf{Learning perceptually-meaningful Depth-based features.}}} \ 
%\subsection{Learning perceptually-meaningful depth features} 
\label{LearningPercept}
%\subsection{Depth-based perceptual features \& perceptual similarity} \label{LearningPercept}
% We used the learned depth-perceptual features in multiple places
%\subsection{Perceptually-meaningful depth features \& Depth-based perceptual similarity} \label{LearningPercept}
%To comply with the self-supervised approach of~\cite{Gaziv2020}
% We extend the concept of Perceptual Similarity~\cite{Zhang2018TheMetric} from images to depth maps. 
% This requires learning perceptually-meaningful  features. 
%To achieve this, 
%we train networks to learn depth perceptual features~(Fig~\ref{fig:Method}b). 
% We used these networks for three purposes: (i)~it provides the main fixed features-extractor network within the Encoder, (ii)~it underlies the perceptual similarity metric, which is used as a reconstruction criterion (of images or depth maps), and (iii)~it is used for metric-based evaluation of the reconstructed depth maps.
%To this end,
Part of the success of our method  stems from 
%learning and 
using perceptually-meaningful Depth-based  features. 
%These learned perceptual Depth features 
These features are used in two main ways within our method: 
(i)~They form the basis for our \emph{Depth-based Perceptual Similarity} metric (explained next);
%, which is used as a reconstruction criterion within our Decoder loss;  
(ii)~They form the main backbone of our Encoder 
%(via a fixed features-extractor network -- 
(a detailed description of our networks architecture is found in the Supplementary-Material).
%
%These learned perceptual deep features 
% These features are used in two main ways: 
% (i)~They form the basis for our \emph{Depth-based Perceptual Similarity} metric, which is used as a reconstruction criterion within our Decoder loss;  
% (ii)~They form the main backbone (via a fixed features-extractor network) within our Encoder -- a detailed description of our networks architecture can be found in the Supplementary-Material.
These perceptually-meaningful depth-based features are learned as follows:
We customized the VGG architecture to accommodate a four-channel RGBD input, and another version that respects a single Depth-only input. We then train these networks from scratch on ImageNet for the task of {object recognition}~(Fig~\ref{fig:Method}b), similar to the way the original VGG network was trained. 
To obtain the ``ground-truth'' depth data required for this training, we estimated the depth maps of all $\sim$1.3M images in ImageNet classification challenge~(ILSVRC)~\cite{Deng2009ImageNet:Database} using MiDaS.
% These data is then used in the following steps. 
We denote the trained two \emph{\underline{depth-based} object recognition networks} as $\varphi_{RGBD}$ and $\varphi_{D}$.
Notably, the resulting object recognition accuracy of $\varphi_{RGBD}$ 
%was insensitive to adding the depth channel to the input, performing 
was comparable with its RGB-only baseline ($\sim$70\% accuracy),
%On the other hand, using the depth channel alone as input  while discarding RGB channels was sufficient to achieve
%over 60\% of the baseline performance ($\sim$42\% accuracy).
whereas the depth-only recognition network $\varphi_{D}$ achieved $\sim$42\% accuracy.
%We use the train networks as the main fixed features-extractor network within the Encoder, as well as the basis for our Depth-based Perceptual Similarity, which we explain next.
The goal, however,  was not to improve object recognition, but rather to obtain a coarse-to-fine hierarchy of semantically-meaningful depth-related features.

\underline{\textit{\textbf{Depth-based Perceptual Similarity.}}}
We extend the concept of Perceptual Similarity metric of~\cite{Zhang2018TheMetric} from images to depth-related data (Depth-only or RGBD). 
%We define a new version of Perceptual Similarity metric, which applies to depth data, using the Depth-only or RGBD trained networks. 
%Specifically, given a reconstructed depth map~$\hat{D}$ and a ground-truth depth map~$D$, 
% we extract features from multiple layers of the trained Depth-only recognition network, $\varphi_D$ (from low to high layers, corresponding to lower-to-higher ``semantic'' levels).
%we use the Depth-only recognition network, $\varphi_D$, to 
Specifically,  to compare a reconstructed depth map $\hat{s}$ to a target depth map $s$, we first feed both into the trained Depth-only recognition network, $\varphi_D$, and extract features from multiple blocks (from low to high layers, corresponding to lower-to-higher ``semantic'' levels).
%The perceptual metric is then computed by feature similarity at corresponding levels. 
%An equivalent the perceptual metric is  defined for
Similarly, when $s$ and $\hat{s}$ are RGBD data, we use the trained RGBD-based recognition network, $\varphi_{RGBD}$. 
For brevity we henceforth refer to both networks as $\varphi$.
%
% We denote the deep features extracted from a Depth or RGBD input, $s$, 
We denote the deep features extracted from an input $s$  
at the output of a particular block $b$, by $\varphi^{b}\left(s\right)$.
The perceptual similarity between $s$ and $\hat{s}$, $\mathcal{L}_{perceptual}\left(\hat{s},s\right)$, is then defined by cosine similarity between channel-normalized ground-truth and predicted features at each block output:
% \vspace*{-0.3cm}
\vspace*{-0.12cm}
\begin{equation}\label{perceptual_similarity}
\mathcal{L}_{perceptual}\left(\hat{s},s\right)\propto-\sum_{b=1}^{5}\cos\left(\angle\left(\varphi^{b}\left(\hat{s}\right),\varphi^{b}\left(s\right)\right)\right), 
% \mathcal{L}_{perceptual}\left(\hat{s},s\right)\propto-\sum_{b=1}^{5}\cos\left(\angle\left(\varphi_{RGBD}^{b}\left(\hat{s}\right),\varphi_{RGBD}^{b}\left(s\right)\right)\right), 
\end{equation}
%
%where $\hat{s}$ and $s$ are the reconstructed and target data, respectively.

% We next describe the integration of the surrogate depth data and the learned depth perceptual features within the self-supervised approach to achieve depth reconstruction from fMRI.

\vspace*{-0.15cm}
%\vspace*{-0.5cm}
\underline{\textit{\textbf{Handling the insufficient fMRI training data - The self-supervised approach:}}}

Fig~\ref{fig:Method}cd shows our two-phase self-supervised training. 
% While we describe and show the method for RGBD encoding/decoding, the details also apply for Depth-only configuration of the framework, but discarding any RGB references.
We describe the method for the case of RGBD encoding/decoding, and then briefly highlight the differences in the Depth-only configuration. 

$\bullet$ \textit{\textbf{Encoder supervised training (Phase I, Fig~\ref{fig:Method}c).}} \  
% \subsubsection*{Encoder supervised training (Phase I)} 
Let $r$ be the ground truth fMRI, and  $\hat{r}=Enc\left(s\right)$ denote the 
encoded fMRI resulting from applying our Encoder to an RGBD input, $s$.
% encoded fMRI response from an RGBD input, $s$. by our Encoder.
%, by Encoder $Enc$. 
We define an fMRI loss by a convex combination of mean square error and cosine proximity between $\hat{r}$ 
and  $r$:
%with respect to the ground truth fMRI, $r$:
%. The \textbf{fMRI loss} is defined as:
%
% \begin{equation} \label{fmri_loss}
% \mathcal{L}_{r}\left(\hat{r},r\right)=\alpha\left\lVert {\normalcolor \hat{r}-r}\right\rVert_{2}-\left(1-\alpha\right)\cos\left(\angle\left(\hat{r},r\right)\right),
% \end{equation}
\begin{equation} \label{fmri_loss}
\mathcal{L}_{r}\left(\hat{r},r\right)=\alpha\cdot\text{MSE}\left(\hat{r}, r\right)-\left(1-\alpha\right)\cos\left(\angle\left(\hat{r},r\right)\right),
\end{equation}
%
% \noindent
where $\alpha$ is a hyperparameter set empirically ($\alpha=0.9$). We use this loss for training the Encoder $E$. 
% \lbreak
% \inred{
% Notably, in the considered fMRI datasets, the subjects who participated in the experiments were instructed to fixate at the center of the images. Nevertheless, eye movements were not recorded during the scans, thus the fixation performance is not known. To account for the center-fixation uncertainty, we introduced small random shifts (+/- a few pixels) of the input images during Encoder training. This resulted in a substantial improvement in the  Encoder performance and subsequently in the RGBD reconstruction quality.
% }
% %
Upon completion of Encoder training, we proceed to training the Decoder with a fixed Encoder. 

% \subsubsection{Decoder training (Phases II, IV)}
% \subsubsection{Decoder training (Phase II)} \label{PhaseII}
$\bullet$ \textit{\textbf{Decoder training (Phase II, Fig~\ref{fig:Method}d).}} \ 
% \subsubsection*{Decoder training (Phase II)} \label{PhaseII}
Decoder training is driven by two main losses: 
\begin{equation}\label{dec_loss}
    % \mathcal{L}^{D}+\mathcal{L}^{ED}+\mathcal{L}^{DE}.
    \mathcal{L}^{Dec}+\mathcal{L}^{EncDec},
\end{equation}
where $\mathcal{L}^{Dec}$ is a supervised loss on training pairs of image-fMRI, and $\mathcal{L}^{EncDec}$ (Encoder-Decoder) is an \emph{unsupervised} loss on unpaired images (without corresponding fMRI recordings). Both components of the loss are normalized to have the same order of magnitude (all in the range $[0, 1]$, with equal weights), to guarantee that the total loss is not dominated by any individual component. We found our reconstruction results to be relatively insensitive to the exact balancing between the two-loss components.
%(see Supplementary-Material).
%
We next detail each component of the loss.

% \lbreak
$\mathcal{L}^{Dec}$: \emph{Decoder Supervised Training} (Fig~\ref{fig:Method}d1). 
Given  training pairs~$\{\left(r, s\right)\}$=\{fMRI, RGBD\}, the supervised loss $\mathcal{L}^{Dec}$ is imposed on the decoded RGBD image, $\hat{s}$$=$$Dec\left(r\right)$. 
%and is defined via an RGBD reconstruction objective, $\mathcal{L}_{s}$, as: 
$\mathcal{L}^{Dec}$$=$$\mathcal{L}_{s}\left(\hat{s},s\right)$
%
%$\mathcal{L}_{s}$ 
consists of an $\ell_{1}$-loss on the RGBD values, 
%$\mathcal{L}_{RGB}$ and $\mathcal{L}_{depth}$, 
as well as our depth-based perceptual loss, $\mathcal{L}_{perceptual}$ (Eq.~\ref{perceptual_similarity}):
\begin{equation}\label{image_loss}
\mathcal{L}_{s}\left(\hat{s},s\right)=
%\mathcal{L}_{RGB}\left(\hat{s}_{rgb},s_{rgb}\right)+\mathcal{L}_{depth}\left(\hat{s}_{dep},s_{dep}\right)
\left\lVert {\normalcolor \hat{s}-s}\right\rVert _{1} 
+\mathcal{L}_{perceptual}\left(\hat{s},s\right)+\mathcal{R}\left(\hat{s}\right)
\end{equation}
where, $\mathcal{R}\left(\hat{s}\right)$, corresponds to total variation (TV) regularization of the reconstructed (decoded)  $\hat{s}$. 
% The depth criterion penalize for the Mean-Average-Error of the predicted map.
% % and of the transitivity map using attenuation coefficient $\beta=2$. 
% Furthermore, additional criteria apply on the predicted depth via the RGBD perceptual loss, $\mathcal{L}_{perceptual}$. 
% The Image loss in Eq.~\ref{image_loss} is also used for the self-supervised Encoder-Decoder training on \texit{unpaired images and corresponding depth maps} (images \& depth maps without fMRI), as explained next.

$\mathcal{L}^{ED}$:  \emph{\textbf{Self-supervised} Encoder-Decoder training on unpaired Natural Images \& Depth Maps} (Fig~\ref{fig:Method}d2). This objective enables to train on any desired unpaired image along with its corresponding depth map (images for which fMRI was never recorded), well beyond the 1200 images included in the fMRI dataset. In particular, we used $\sim$50K additional natural images from ImageNet's 1000-class data~\cite{Deng2009ImageNet:Database}, along with their estimated depth-maps (see Sec.~\ref{sec:approach}). 
%\emph{This allows adaptation to the statistics of many more novel semantic categories, thus learning the common higher-level feature representation of various novel classes.}
% To train on RGBD data without corresponding fMRI responses, we map images to themselves through our Encoder-Decoder transformation,
We train on such RGBD data without fMRI, by imposing cycle consistency through our Encoder-Decoder transformation:
\begin{equation*}
    s\mapsto\hat{s}_{EncDec}=D\left(E\left(s\right)\right).
\end{equation*}
The unsupervised component $\mathcal{L}^{EncDec}$ of the loss in Eq.~\ref{dec_loss} on unpaired images, $s$, reads:
\begin{equation*}
    \mathcal{L}^{EncDec}=\mathcal{L}_{s}\left(\hat{s}_{EncDec},s\right),
\end{equation*}
where $\mathcal{L}_{s}$ is the Image loss defined in Eq.~\ref{image_loss}. 
In other words, $\mathcal{L}^{EncDec}$ imposes \underline{cycle-consistency} on any RGBD data, but at a \underline{perceptual level} (not only at the pixel level). 
%Note that this high-level perceptual consistency does \underline{not} require using any class labels in the training.

The input to our Encoder (and output of our Decoder) are 112$\times$112 images and depth maps, although our method works well also on other resolutions. For details on hyperparameters as well as on Encoder/Decoder architectures see Supplementary-Material.

\textit{\textbf{Depth-only framework.}} \ 
In a Depth-only configuration we encode/decode (reconstruct) the depth map alone, discarding all RGB data from training. Specifically, 
%we set $\mathcal{L}_{RGB}=0$, 
we switch 
%the perceptual loss, $\varphi_{RGBD}^{b}$, 
to a Depth-only perceptual loss using $\varphi_{D}^{b}$. The Encoder/Decoder switch to a single channel input/output, respectively. Particularly, the Encoder architecture switches to using a Depth-only pretrained network.

\textit{\textbf{Runtime.}} \ Our system completes the two-stage training within approximately 2 hours on a single Tesla V100 GPU. Inference (decoding of new fMRI recordings) takes a few milliseconds per image.
% We successfully applied our method to fMRI datasets as follows.

\vspace*{-0.2cm}
\section{Experiments and Results}
\label{sec:experiments}
\vspace*{-0.2cm}
% \vspace*{-0.25cm}

% \textit{\textbf{Experimental datasets.}} \ 
\subsection{Experimental datasets} \label{ExperimentalDatasets}
\vspace*{-0.1cm}
We tested our self-supervised depth reconstruction approach on a highly popular and publicly available benchmark fMRI dataset, called \href{https://openneuro.org/datasets/ds001246/versions/1.0.1}{\color{magenta}{fMRI on ImageNet}}~\cite{Horikawa2015GenericFeatures}. We found this dataset to be the only one currently suitable for our method in terms of the stimuli used and fMRI signal quality. An important point is the requirement to perform depth estimation on color ImageNet-like natural images that match the input distribution expected by the MiDaS depth estimation network we used. 
% However we look forward to applying our method to the upcoming `Natural Scene Dataset'~\cite{Allen2020TheStudy}, which appears to be ideal for the task.

% `fMRI on ImageNet' provides fMRI recordings paired with their corresponding underlying images. Subjects were instructed to fixate on a cross positioned at the center of the presented images. \textbf{`fMRI on ImageNet'} comprises 1250 distinct ImageNet images drawn from 200 selected categories. The train- and test-fMRI data consist of 1 and 35 {(repeated recordings)} per presented stimulus, respectively. Fifty image categories provided the fifty test images, one from each category. The remaining 1200 were defined as train set (with only one fMRI recording). The images in the train set and test set come from mutually exclusive categories (different classes). We considered approximately 4500 voxels from the visual cortex provided by the authors of~\cite{Horikawa2015GenericFeatures}. 
% %
% The images were equally delivered to both of the subject's eyes with no binocular disparity.

`fMRI on ImageNet' provides fMRI recordings of observed images. Subjects were instructed to fixate on a cross positioned at the center of the presented images. \textbf{`fMRI on ImageNet'} comprises 1250 distinct ImageNet images drawn from 200 selected categories. 
%The train- and test-fMRI data consist of 1 and 35 {(repeated recordings)} per presented stimulus, respectively. 
Fifty image categories provided the 50 test images, one from each category. The remaining 1200 were defined as train set.
% (with only one fMRI recording). 
The images in the train set and test set come from mutually exclusive categories (different classes). We considered approximately 4500 voxels from the visual cortex, provided by the authors of~\cite{Horikawa2015GenericFeatures}. 
%
%The images were equally delivered to both of the subject's eyes with no binocular disparity.

We used additional $\sim$50K \textbf{\emph{``unpaired'' natural images from ImageNet}}'s validation set~\cite{Deng2009ImageNet:Database}  \textbf{ \emph{with their estimated depth maps}}, for our self-supervised training (Fig.~\ref{fig:Method}d2).
%on unpaired RGBD data (Encoder-Decoder objective, Fig.~\ref{fig:Method}d2). 
We verified that the images in our additional unlabeled external dataset, are distinct from those in the `fMRI on ImageNet'.
% Michal - there is some 20-class overlap of classes between the test and the external images, remember? but not interesting here.

% \vspace*{-0.1cm}
\vspace*{-0.12cm}
% \vspace*{-0.2cm}
\subsection{Depth recovery - results \& evaluations}
\vspace*{-0.1cm}
Fig.~\ref{fig:MainResults} shows visual results of our proposed self-supervised RGBD method. This includes the reconstructed RGB images (of the underlying images seen by the subjects) and their corresponding reconstructed depth maps (never seen by the subjects) -- both recovered directly from fMRI. 
%(Fig \ref{fig:TaskMethod}b,d,e). 
%
A stark feature that emerges in the reconstructed depth maps is foreground/background segregation. However, a closer inspection reveals finer details of depth variations. 
% <For example XX>
%
% Interestingly, some reconstructed depth maps even reveal depth information that is superior to its ``ground-truth'' counterpart (i.e., the depth map estimated from the original RGB-image). <For example XX>.
%
% Importantly, the presented results demonstrate strong generalization, showing results of the entire test cohort (no cherry-picking). These results correspond to Subject 3, with the highest noise-ceiling fMRI data (see Supplementary-Material for Subjects 1,2,4, and 5).
Results are shown for the 
entire test cohort (all 50 test fMRIs of the `fMRI on ImageNet' dataset).
% test cohort of the `fMRI on ImageNet' dataset (for results of all 50 test fMRIs see Supplementary-Material). 
These results correspond to Subject 3, who has the highest noise-ceiling fMRI data (results on other subjects are in the Supplementary-Material).

\begin{SCfigure}
%\vspace*{-0.5cm}
% \hspace*{-4cm}
\hspace*{-1cm}
\includegraphics[scale=0.65]{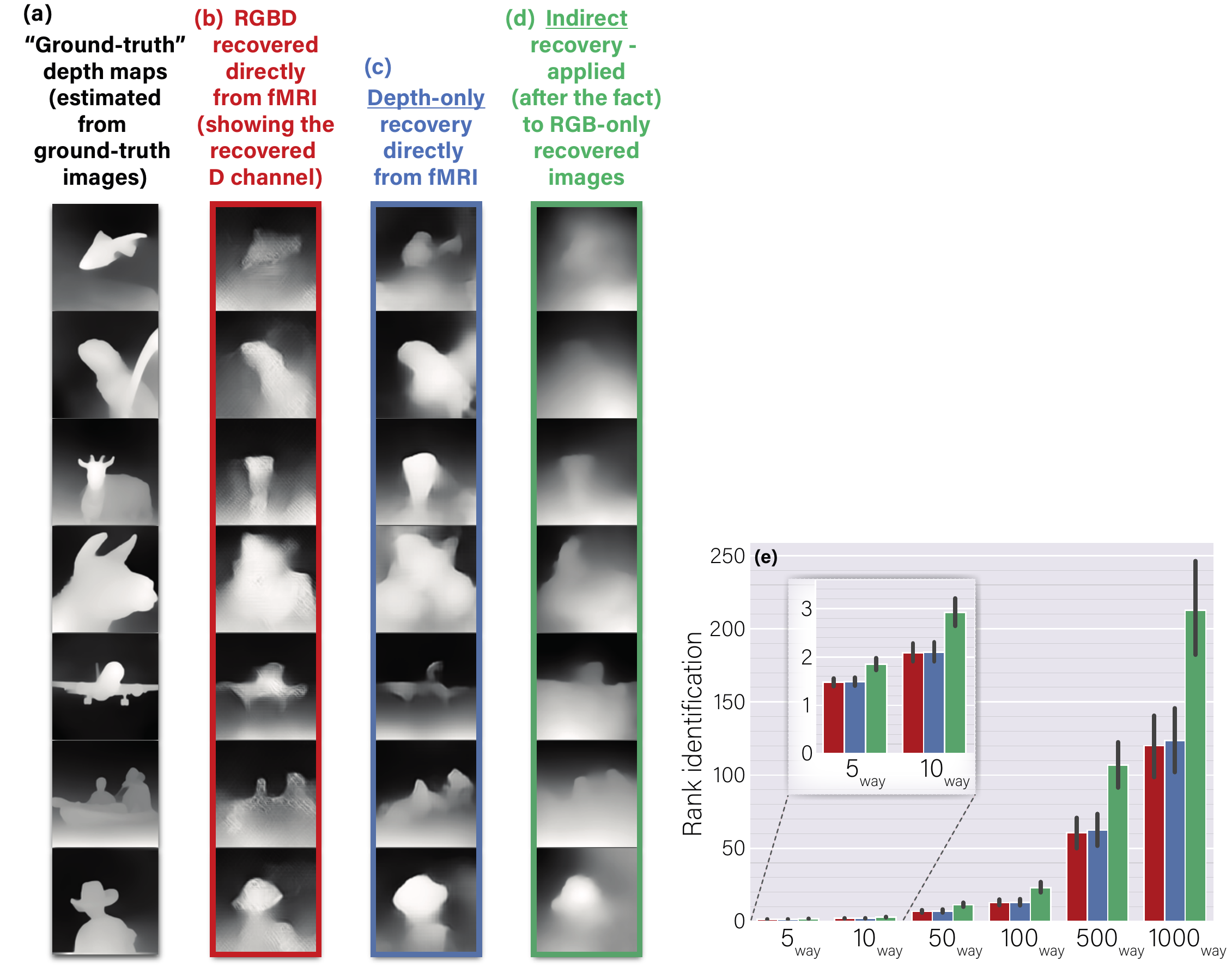}
\hspace*{-7cm}
% \hspace*{-6.5cm}
% \hspace*{-4cm}
\caption{
% \textbf{Depth is best recovered directly from fMRI (visual \& quantitative).} \
\textbf{Depth recovery by RGBD \& Depth-only approaches.} \
{\small {\it \textbf{(a)}~Depth map estimated from ground-truth RGB-image using MiDaS~\cite{Lasinger2019TowardsTransfer}. \textbf{(b)}~Depth channel of RGBD reconstruction directly from fMRI. \textbf{(c)}~Depth recovery directly from fMRI, where both Encoder \& Decoder are trained ``Depth-only'' -- without using any RGB data for training. \textbf{(d)}~Indirect depth recovery from reconstructed RGB-images using MiDaS. This approach fails to recover depth faithfully.  \textbf{(e)}~Depth reconstruction mean rank for \textbf{(b)}-\textbf{(d)} respectively by n-way rank identification experiments (lower is better, showing average over all 50 reconstructed depth maps and five subjects). $95\%$ Confidence Intervals by bootstrap shown on charts.  
% <CHANGE FIGURE TO SELECTED IMAGES XX>
}}}
\label{fig:DepthRecoveryOptions}
% \end{figure}
% \vspace*{-0.5cm}
\vspace*{-0.9cm}
\end{SCfigure}

\textit{\textbf{Performance evaluation.}} \ 
% \subsection{Performance evaluation} 
% To quantify the quality of the reconstructed images and their depth maps, we followed 
To quantitatively evaluate our reconstruction results, we followed 
%a retrieval approach in 
an n-way identification experiment~\cite{Gaziv2020, Beliy2019FromFMRI, Shen2019DeepActivity, Ren2021ReconstructingLearning, Seeliger2018GenerativeActivity, Zhang2018Constraint-FreeNetwork}, applied  separately to images (RGB) and to depth maps (D).
%
%In this approach, 
Each reconstructed image is compared against $n$ candidate images (the ground truth image, and $(n-1)$ other randomly selected images). The goal is to identify the ground truth, or at least rank it well  among the candidates (rank=1 signifies perfect identification). 
We evaluate our RGB or Depth reconstructions using this \underline{rank} identification (lower is better). This provides an informative accuracy measure that accounts for cases when the ground truth is not strictly identified as the best candidate, but is nevertheless ranked fairly low.
Reconstructed \textit{Images (RGB)} were compared against candidates using the image-based Perceptual Similarity metric of~\cite{Zhang2018TheMetric}.
%as a distance metric~\cite{Gaziv2020}.
%
%Analogously, we compare
Reconstructed \emph{depth maps} were compared against candidate depth maps using our new \emph{Depth-only} Perceptual Similarity metric.

Fig~\ref{fig:DepthRecoveryOptions} shows comparison of our three main approaches for depth recovery (see Sec~\ref{ApproachOverview}).
Specifically, it shows results for our two main approaches to depth-decoding from fMRI, which achieved the best results: (i)~RGBD framework (Fig~\ref{fig:DepthRecoveryOptions}b), and (ii)~Depth-only framework (training for depth recovery without any RGB data during training, Fig~\ref{fig:DepthRecoveryOptions}c).
We find that both approaches successfully recover depth details as demonstrated in the visual results. Furthermore, quantitative rank identification evaluation scores a mean rank of 120 and 124 in the challenging 1000-way task for RGBD and Depth-only methods, respectively -- more than 4x the chancel level (chance level rank=500, showing average scores over all 5 subjects). Comparing these approaches, we find that the results by both approaches are largely on par with each other. This implies that best depth recovery results are obtainable even in \textit{complete absence of any RGB supervision or self-supervision}. On the other hand, the RGBD approach provides additional recovered information -- the reconstructed RGB image.

% following a Depth-only approach gave rise to comparable depth recovery results to those obtained by the RGBD approach. This is despite 

% We next analyzed whether RGBD-based depth recovery is dominantly driven by the reconstructed RGB component. To this end, we trained a depth-only framework (Encoder and Decoder), where we completely discarded all RGB data during training\footnote{To align with a depth-only framework, all the pretrained networks used were changed to receive as input a single depth-only channel. This holds for the backbone network within the Encoder architecture and in the Perceptual Similarity loss.}. We then used the trained depth-only Decoder to reconstruct the test depth-maps, directly from fMRI (Fig~\ref{fig:DepthRecoveryOptions}d). We find that the results by this approach are largely on a par with those by direct RGBD reconstruction, even under \textit{complete absence of the RGB supervision or self-supervision}. 

% \vspace*{-0.1cm}
\vspace*{-0.25cm}
\subsection{Ablation study – The importance of Direct vs. Indirect depth reconstruction from fMRI} \label{sec:ablations}
% \textit{\textbf{Ablation study – The importance of Direct vs. Indirect depth reconstruction from fMRI.}} \\
\vspace*{-0.25cm}
We compare our reconstruction results agaist several baselines:

$\bullet$ \textit{\textbf{RGB-only Enc/Dec} \underline{followed} by depth estimation on the reconstructed images.}  
Fig~\ref{fig:DepthRecoveryOptions}d shows results for an \underline{indirect} depth recovery approach. We estimate the depth map from a reconstructed RGB image as a post-processing step (The RGB image is reconstructed via an RGB-only framework with the perceptual features/similarity of~\cite{Zhang2018TheMetric}).
%, see Sec~\ref{ApproachOverview}). 
This involves no training on depth data at all. This approach gives rise to poor depth reconstruction quality, indicated both visually and quantitatively in Fig~\ref{fig:DepthRecoveryOptions}.

$\bullet$ \textit{\textbf{RGB-only Enc/Dec}, but  \underline{constrained} by a depth estimation loss on the reconstructed images.} \ 
% The indirect depth recovery approach previously mentioned uses MiDaS depth estimation to recover the depth maps from reconstructed images. However, these 
We extended the \emph{indirect} depth recovery of the RGB-only framework,
%by adding MiDaS depth estimation on the reconstructed image \underline{during training}. 
by imposing a depth loss on the reconstructed RGB images \underline{during training}.
To impose depth reconstruction criteria on an RGB-only Decoder, we mounted atop it the pre-trained MiDaS depth estimation network~\cite{Lasinger2019TowardsTransfer}, $\mathcal{M}\left(\cdot\right)$. Although the Decoder itself does not produce any depth map, combining the resulting MiDaS depth map with the decoder's RGB output provides an RGBD output in total. We experimented with two types of losses: (i) the standard loss of the RGBD framework, $\mathcal{L}_{s}$, which includes also depth perceptual similarity $\mathcal{L}_{perceptual}$ (Eq.~\ref{perceptual_similarity}), and (ii) a simple $\ell_{1}$-loss on the depth maps, $\left\lVert {\normalcolor \mathcal{M}\left(\hat{s}_{rgb}\right)-\mathcal{M}\left(s_{rgb}\right)}\right\rVert _{1}$. 

\begin{figure} 
% \begin{SCfigure}
% \vspace*{-.5cm}
\vspace*{-.3cm}
\hspace*{-1.2cm}
% \begin{SCfigure}
    % \includegraphics[scale=\figscaleC]{StarterResults2.pdf}
% \includegraphics[scale=\figscaleA]{GeneralAblations.pdf}
\includegraphics[scale=0.55]{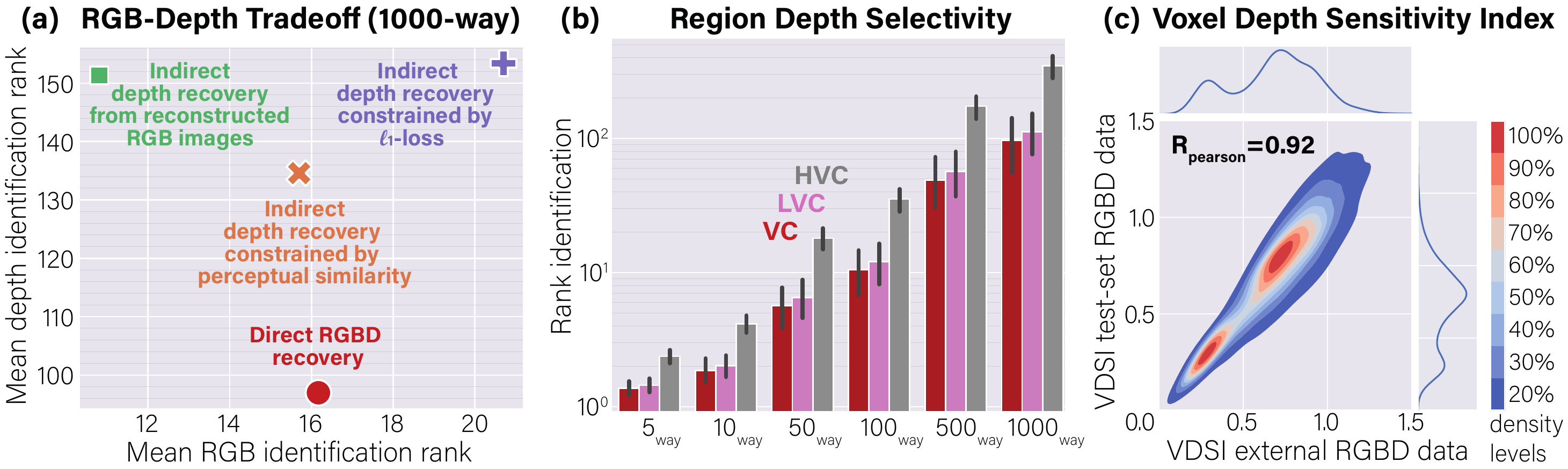}
% \hspace*{-5cm}
% \vspace*{-0.7cm}
\vspace*{-0.5cm}
\caption{
\textbf{Ablations study.} 
{\small {\it 
(See Sec~\ref{sec:ablations} for details). 
% Showing for Subject 3; see Supp.-Material for other subjects.
% Showing for Subject 3 (see Supplementary-Material for other subjects).
}}
% {\small {\it \textbf{(a)}~RGB-Depth reconstruction trade-off. We evaluated the RGB and depth reconstruction quality by rank identification (lower is better). Our proposed 
% Direct RGBD recovery approach provides the best depth reconstruction. \textbf{(b)}~Rank identification evaluations of reconstructed depth maps when using only a subset of voxels from the Lower Visual Cortex only (LVC), the Higher Visual Cortex (HVC), or all voxels from the Visual Cortex. Only voxels from the selected regions are used for both training and inference. Depth reconstruction quality is driven by voxels from LVC. \textbf{(c)}~Voxel Depth Sensitivity Index (VDSI), when computing index values using RGBD data from fMRI test set or the external dataset. We show the Density (KDE) resulting from the two index values per voxel. We find strong agreement across different sources of RGBD data used. Most voxels have VDSI$<1$, thus jointly tuned to RGB and depth information. Results shown for Subject 3. $95\%$ Confidence Intervals by bootstrap shown on charts. }}
%     {\small {\it 
%     \textbf{(Left column):} Depth maps reconstruction results side-by-side their estimated ground-truth from the original RGB images. 
%     \textbf{(Right column):} Image (RGB) reconstruction results side-by-side the ground-truth images presented to the human. 
%     We show results of the entire test cohort of \mbox{\href{https://openneuro.org/datasets/ds001246/versions/1.0.1}{\color{magenta}{fMRI on ImageNet}}~\cite{Horikawa2015GenericFeatures}}. 
%     }}
    }
    \label{fig:GeneralAblations}
    % \vspace*{-0.3cm}
    % \vspace*{-.5cm}
    \vspace*{-.53cm}
    % \vspace*{-0.585cm}
\end{figure}

%Fig~\ref{fig:GeneralAblations}a plots the mean \emph{depth-rank}  versus the mean \emph{RGB-rank}  
The y-axis in Fig~\ref{fig:GeneralAblations}a shows the mean \emph{depth-rank}
in the challenging \emph{1000-way} identification task  
(lower is better; shown for Subject 3; see Supp-Material for all subjects). 
The direct RGBD approach 
%for depth recovery 
scores mean \emph{depth-rank} of 97 (5x better than chance level), significantly outperforming the indirect approaches,  (i) \& (ii), by a large margin (38 and 56 rank levels, respectively). The depth recovery performance achieved by the  \emph{perceptually-constrained} indirect approach is  slightly better than  the two other indirect approaches for depth recovery (but significantly worse than the direct ones).
%vanilla approach of depth recovery from reconstructed RGB images as a post-process step.

% We repeated (i) using either and RGB-only Encoder (used for E-D training) or an RGB\textit{D} Encoder.

$\bullet$ \textit{\textbf{The RGB-Depth reconstruction Trade-off.}} \ 
% (this is the first time I would point to the graph with RGB rank results, and then the tradeoff graph).}}
Fig~\ref{fig:GeneralAblations}a further plots the mean \emph{depth-rank}  versus the mean \emph{RGB-rank}  in the challenging \emph{1000-way}  identification task, for all the direct \& indirect image+depth recovery approaches.
%
% We studied the interplay between RGB and Depth reconstruction. Fig\ref{fig:GeneralAblations}a shows on a 2D-plane the evaluated RGB and Depth rank identification for different approaches in a 1000-way identification task. These methods include (i)~our direct RGBD approach, (ii)~indirect post-process depth-recovery on reconstructed images, and (iii)~indirect approaches with imposed depth reconstruction criteria.
%
Our results show that our direct RGBD approach provides the best depth reconstructions with yet a very good RGB identification score (a mean-RGB rank of 16 -- more than 30x better than chance level). 
%This score is comparable or slightly better than the indirect approaches with imposed depth recovery criteria.
On the other hand, the RGB-only approach, which is trained solely for RGB recovery, gives rise to better (yet comparable) RGB reconstructions. 
% In summary, It is also comparable or just  with the  suggests no significant compromise to RGB recovery. On the other there exist other methods which provide good depth reconstruction results (e.g., midasdec XX) while compromising the RGB reconstruction quality, and vice versa (e.g., midas1 XX).

% \end{SCfigure}
%

% \vspace*{-0.1cm}
\vspace*{-0.12cm}
\subsection{Detecting depth-sensitive voxels}
\vspace*{-0.1cm}
\emph{\textbf{Predominance of Lower Visual Cortex.}} \ We analyzed the impact of using only a subset of voxels from particular brain regions. Fig~\ref{fig:GeneralAblations}b shows rank identification results when using only voxels from the Lower Visual Cortex (LVC, comprising V1-V3) or those from the Higher Visual Cortex 
% (HVC, comprising the Lateral Occipital Cortex, Fusiform Face Area, and Parahippocampal Place Area). 
(HVC, comprising the LOC, FFA,  PPA). 
In those experiments  the Encoder and  Decoder were trained on the subset of voxels using our RGBD method. We then evaluated the reconstructed depth maps by n-way rank identification ($n$=5, 10, 50, 100, 500, or 1000, lower is better). For comparison, we also plot the results when using all Visual Cortex voxels. We find that our depth reconstruction performance is dominantly driven by the activations from LVC voxels. Using voxels from HVC alone significantly degrades our depth recovery performance.

\textit{\textbf{Voxel-specific Depth Sensitivity Index.}} \ 
% (discuss the voxel-specific selectivity)
We studied whether voxels can be characterized by their tuning to depth versus RGB-image information type. 
% We studied whether voxels can be characterized by their tuning to depth information versus their tuning to RGB-image information. 
To this end, we used our trained RGBD Encoder, and computed the predicted voxel activations 
when setting each one of the RGBD input channels to zero 
%(either the $D$ channel, or a color channel $c$ -- 
(one channel at a time).
%
% for 5 types of RGB-image and depth map pairings. These include response to the regular RGBD inputs, response to original RGB inputs with zeros as depth channel, and response to zero inputs in either of the R, G, or B channels and with the ``original'' depth maps (which were estimated from the ground-truth RGB-images). 
More formally, given a set of  RGBD images, $\{(s_i)\}$, 
%$\{(s_{RGBD}^{i})\}_{i}$, 
we denote by $s_{ki}$ the resulting image when setting channel $k$ in $s_i$ to zero ($k\in\{R,G,B,D\}$). $F_i=Enc(s_i)$ denotes the encoded fMRI of the original RGBD image $s_i$,  $F_{ki}=Enc(s_{ki})$ denotes the encoded fMRI after zeroing channel $k$ in $s_i$.
We  define Voxel Depth Sensitivity Index (VDSI) at voxel $v$ as:
%
% Formally, given an RGBD dataset, 
% $(s_{RGB}^{i}, s_{depth}^{i})_{i}$, 
% we computed the fMRI activation of voxel $v$ for $(s_{RGB}^{i}, s_{depth}^{i})_{i}$, $(s_{RGB}^{i}, 0)_{i}$, $(s_{R=0,GB}^{i}, s_{depth}^{i})_{i}$, $(s_{R,G=0,B}^{i}, s_{depth}^{i})_{i}$, and  $(s_{RG,B=0}^{i}, s_{depth}^{i})_{i}$. The computed fMRI is denoted by $F_{ki}^{v},\;k\in\{1,2,3,4,5\}$, where $F_{ki}^{v}=Enc(s_{ki})$. We then define Voxel Depth Sensitivity Index (VDSI) as,
%
% $\left(s_{RGB}^{i}, s_{depth}^{i}\right)_{i}$, 
% we computed the fMRI activation of voxel $v$ for $\left(s_{RGB}^{i}, s_{depth}^{i}\right)_{i}$, $\left(s_{RGB}^{i}, 0\right)_{i}$, $\left(s_{R=0,GB}^{i}, s_{depth}^{i}\right)_{i}$, $\left(s_{R,G=0,B}^{i}, s_{depth}^{i}\right)_{i}$, and  $\left(s_{RG,B=0}^{i}, s_{depth}^{i}\right)_{i}$. The computed fMRI is denoted by $F_{ki}^{v},\;k\in\left\{1,2,3,4,5\right\}$, where $F_{ki}^{v}=Enc\left(s_{ki}\right)$. We then define Voxel Depth Sensitivity Index (VDSI) as,
% \vspace*{-0.3cm}
% \vspace*{-0.1cm}
\begin{equation} \label{VDSI}
% \frac{\mathbb{E}_{imgs}\left(\left|F_{rgbd}-F_{rgb,d_{zero}}\right|\right)}{\mathbb{E}_{c}\left(\mathbb{E}_{imgs}\left(\left|F_{rgbd}-F_{rgb_{zero\;c},d}\right|\right)\right)}
%
% \text{VDSI}=\frac{\mathbb{E}_{i}\left(\left|F_{1,i}-F_{2,i}\right|\right)}{\mathbb{E}_{c}\left(\mathbb{E}_{j}\left(\left|F_{1,i}-F_{2+c,i}\right|\right)\right)}\;;\;\text{VDSI}\in\left[0,\inf\right).
\text{VDSI}(v)=\frac{\mathbb{E}_{i}\left(\left|F_i(v)-F_{D,i}(v)\right|\right)}{\mathbb{E}_{c}\left(\mathbb{E}_{j}\left(\left|F_i(v)-F_{c,i}(v)\right|\right)\right)}\;;\;\text{VDSI}(v)\in\left[0,\inf\right).
\end{equation}

% \vspace*{-0.23cm}
\vspace*{-0.1cm}
where $c$ is a color channel. Voxels with $\text{VDSI}$$\approx$$1$ have similar sensitivity to color and depth variations. 
%
% Fig~\ref{fig:GeneralAblations}c shows the resulting VDSI values for all Visual Cortex voxels. Specifically, we compute the VDSI using either the RGBD data from the external dataset, or using the RGBD data from the fMRI test set (see Sec~\ref{ExperimentalDatasets}). We then use the two index values computed per each voxel to plot a 2D density of a scatter plot (Kernel-Density-Estimate plot). We found strong agreement of the index across these two conditions (0.92 Pearson's correlation). This agreement was more generally present across using input RGBD data from any two of the three sets: the train and validation from the fMRI dataset, and the external dataset. This consistency suggests the reliability of our index of choice. However, for the vast majority of voxels the index is well below 1, suggesting no particular depth-tuning that surpasses RGB-tuning for almost all of the voxels. 
%
Fig~\ref{fig:GeneralAblations}c shows the resulting VDSI values for all Visual Cortex voxels. 
We evaluated the VDSI on two datasets: (i)~RGBD images from the external dataset, (ii)~the fMRI test set (see Sec~\ref{ExperimentalDatasets}). 
%We use the two index values computed per each voxel to plot a 2D density of a 
The scatter plot (Kernel-Density-Estimate plot) in Fig~\ref{fig:GeneralAblations}c shows a strong agreement of the index across these two datasets (0.92 Pearson's correlation). 
%This agreement was more generally present across using input RGBD data from any two of the three sets: 
Similar agreement was also found with respect to the train and validation sets from the fMRI dataset. This consistency possibly suggests the reliability of our VDSI measure.
%However, for the vast majority of voxels have VDSI   well below 1, suggesting no particular depth-tuning that surpasses RGB-tuning for almost all of the voxels. 
Note, however, that the vast majority of voxels have VDSI  value well below 1. 
%
% \inred{
% Exploring this further and validating the depth-sensitivity of individual brain-voxels  is part of our future work.
Exploring this further and validating the depth-sensitivity of individual voxels is part of our future work.
\vspace*{-0.2cm}
% \vspace*{-1.2cm}
% \vspace*{-0.1cm}
\section*{Conclusion}
\vspace*{-0.2cm}
% \vspace*{-0.25cm}

The proposed method is the first to reconstruct dense 3D depth information from brain activity.
Our approach is capable of  reconstructing depth-based information (\emph{Depth-only} or \emph{RGBD}) directly from fMRI recordings. We compensate for the lack in available fMRI training data by adding self-supervision on a very large collection of  natural images \emph{without fMRI}, along with their depth maps.
We show that predicting the depth map \textit{directly from fMRI} outperforms its indirect recovery from a reconstructed image. 
We further present a Depth-based Perceptual Similarity metric, which employs learned perceptual depth-based features. 
% which is based on learned perceptual depth-based features.
Lastly, we attempt to characterize the degree of depth-sensitivity of brain-voxels via a proposed depth-sensitivity measure. 
Exploring the validity of these depth-sensitivity predictions is part of our future work.
%Our future work will further explore this direction, and will try to validate (or refute) these so-far theoretical findings.
%

% Furthermore, we show that adding depth recovery criteria gives rise also to better image (RGB) reconstruction and results than those obtained without it.
% We also define the Voxel Depth Sensitivity Index as a measure for finding depth-tuned voxels, and demonstrate its validity. We found that the Visual Cortex voxels were largely found not to be ``purely depth-sensitive'' under this measure. This suggests a mixed encoding of RGB- and depth-information within each voxel. 

% \vspace*{-0.2cm}
\section*{Acknowledgments}
% \textbf{Acknowledgments.}
This project has received funding from the European Research Council (\textbf{ERC}) under the European Union’s Horizon 2020 research and innovation programme (grant agreement No 788535).

% \subsection*{Author Contributions}
% \textbf{Author Contributions.} 
% R.B. and G.G. designed the experiments. R.B. implemented the network and conducted the image-reconstruction experiments. G.G. designed and wrote the paper, and analyzed the fMRI data. A.H. conducted reconstruction quality analyses. F.S. and T.G. provided guidance on fMRI preprocessing. M.I. conceived the original idea and supervised the project. All authors discussed the results and commented on the manuscript. 

% Use unnumbered third level headings for the acknowledgments. All acknowledgments
% go at the end of the paper. Do not include acknowledgments in the anonymized
% submission, only in the final paper.

% \medskip
% \clearpage
{\small
\bibliographystyle{ieeetr}
% \bibliography{references}

}

\clearpage

\end{document}